\documentclass[10pt, conference, compsocconf]{IEEEtran}
\usepackage{cite}
\usepackage{graphicx}
\usepackage{float}
\usepackage{amsmath}
\usepackage{listings}
\usepackage{xcolor}
\usepackage{subfig}
\usepackage{url}
\usepackage{commath}
\usepackage{textcomp}
\usepackage{enumitem}

\usepackage{lipsum}

\title{SD-Measure: A Social Distancing Detector}
\makeatletter
\newcommand{\linebreakand}{%
  \end{@IEEEauthorhalign}
  \hfill\mbox{}\par
  \mbox{}\hfill\begin{@IEEEauthorhalign}
}
\makeatother

\author{\IEEEauthorblockN{
Savyasachi Gupta\IEEEauthorrefmark{1},
Rudraksh Kapil\IEEEauthorrefmark{2},
\\
Goutham Kanahasabai\IEEEauthorrefmark{3},
Shreyas Srinivas Joshi\IEEEauthorrefmark{4},
Aniruddha Srinivas Joshi\IEEEauthorrefmark{5}}
\IEEEauthorblockA{B.Tech., Department of Computer Science and Engineering\\
National Institute of Technology, Warangal, Telangana, India - 506004\\
gsavya10@gmail.com\IEEEauthorrefmark{1},
rkapil@student.nitw.ac.in\IEEEauthorrefmark{2},\\
gauthamkanags@gmail.com\IEEEauthorrefmark{3}, sj841871@student.nitw.ac.in\IEEEauthorrefmark{4}, aniruddha980@gmail.com\IEEEauthorrefmark{5}}
}
\begin{document}
\maketitle
\begin{abstract}
The practice of social distancing is imperative to curbing the spread of contagious diseases and has been globally adopted as a non-pharmaceutical prevention measure during the COVID-19 pandemic. This work proposes a novel framework named SD-Measure for detecting social distancing from video footages. The proposed framework leverages the Mask R-CNN deep neural network to detect people in a video frame. To consistently identify whether social distancing is practiced during the interaction between people, a centroid tracking algorithm is utilised to track the subjects over the course of the footage. With the aid of authentic algorithms for approximating the distance of people from the camera and between themselves, we determine whether the social distancing guidelines are being adhered to. The framework attained a high accuracy value in conjunction with a low false alarm rate when tested on Custom Video Footage Dataset (CVFD) and Custom Personal Images Dataset (CPID), where it manifested its effectiveness in determining whether social distancing guidelines were practiced.
\end{abstract}\begin{IEEEkeywords}
Social distancing, Person detection, Mask R-CNN, Centroid-based Object Tracking, COVID-19
\end{IEEEkeywords}
%
\section{Introduction}
\label{intro}
The onset of the COVID-19 pandemic has led to an increase in the importance of social distancing to intervene in the spread of the virus by curbing social interactions and maintaining physical distance. Social distancing can be defined as a non-pharmaceutical disease prevention and control intervention enforced to curb contact between those who are infected with a disease and those who are not, so as to stop or diminish the rate and extent of the transmission of the disease within a community. Eventually, this leads to a decrease in the spread of the disease and the fatalities caused by it. Centers for Disease Control and Prevention (CDC) safety guidelines dictate that a distance of at least 6 feet must be maintained between two individuals in both indoor and outdoor spaces \cite{mref2}.\par 

Common social distancing measures include the closure of schools, universities, non-essential work-spaces, and public transportation. Researchers state that the rapid spread of COVID-19 is likely due to the movement of people with little to no symptoms \cite{mref9}, namely those who are unaware that they even contracted the virus. This is why the researchers have stressed upon the fact that social distancing is such a paramount containment measure. \par

Based on an Indian Council of Medical Research (ICMR) study \cite{mref8}, health officials have elucidated that the current $R_{o}$ or R naught for the coronavirus infection is somewhere between 1.5 and 4, where R naught is a contagious virus’ basic reproductive number. If we take the $R_{o}$ value to be 2.5, then a single infected patient can spread the disease to 406 other humans in roughly 30 days if appropriate social distancing norms are not followed. However, if social exposure is decreased by about 75\% then a single infected patient will only be able to infect 2.5 people on average \cite{mref8}. This emphasises how imperative social distancing can be in curbing the transmission of contagious agents. \par 

Since the imposition of lockdowns and public restrictions in many parts of the world, the task of identifying whether a safe distance that complies with social distancing norms is presented as a new area of study in the field of computer vision. However, it is difficult for humans to estimate the distance between two people accurately from video footage alone, considering that cameras are usually angled in such a way as to maximise the field of vision. As a result, the proportions of objects in the images are distorted and make estimation inaccurate.\par

In this work, we propose a novel approach that allows a system to automatically detect when people are standing too close to each other, and to output an accurate estimate of the distance between them as well
%
\section{Related works and Literature}
\label{relatedworks}
In this section, we review some of the related methodologies that aim to tackle the same problem as our proposed approach. As social distancing has been in the limelight of being practiced in the recent months, there is a dearth of published work in this field. In \cite{mref1}, the author proposes an approach to estimating the distance between people to analyze whether social distancing is maintained. After obtaining the bounding box of people using YOLOv3, a width threshold is set for objects among which the distance is measured. By measuring the ratio of pixels to metres, the distance between two people in a given frame is approximated. However, this approach only calculates the distance between people without taking the individual distances of each person from camera into account. The major assumption is that each person is the same distance from the camera. \par
In a similar approach \cite{mref10}, the author uses the YOLOv3 object detection framework to identify people in a given frame. After computing a pairwise distance between the centroids of the detected bounding boxes of people, this value is compared to a predefined minimum pixel threshold. By mapping these pixels to measurable units, detection of violations in social distancing norms are identified. \par
Punn \emph{et al.} \cite{mref7} proposed a methodology for monitoring COVID-19 social distancing with person detection and tracking. The framework utilizes the YOLOv3 object detection model to segregate people from the background region. A deepsort technique was used to track the identified people with the help of bounding boxes and assigned tracking IDs. After computing the pairwise vectorized L2 norm from the 3-D space obtained from the centroid coordinates and the dimensions of the bounding boxes, a violation index is proposed to identify the infringement of social distancing protocols. However, this approach doesn't measure the distance between each pair of people. As the authors themselves acknowledge, it could have achieved better precision if they had used Faster R-CNN instead of YOLOv3.
%
\section{Proposed Approach}
\label{approach}
\begin{figure*}[!ht]
\centering
\begin{minipage}{1\textwidth}
    \centering
    \includegraphics[width=0.99\linewidth, height=0.42\textheight]{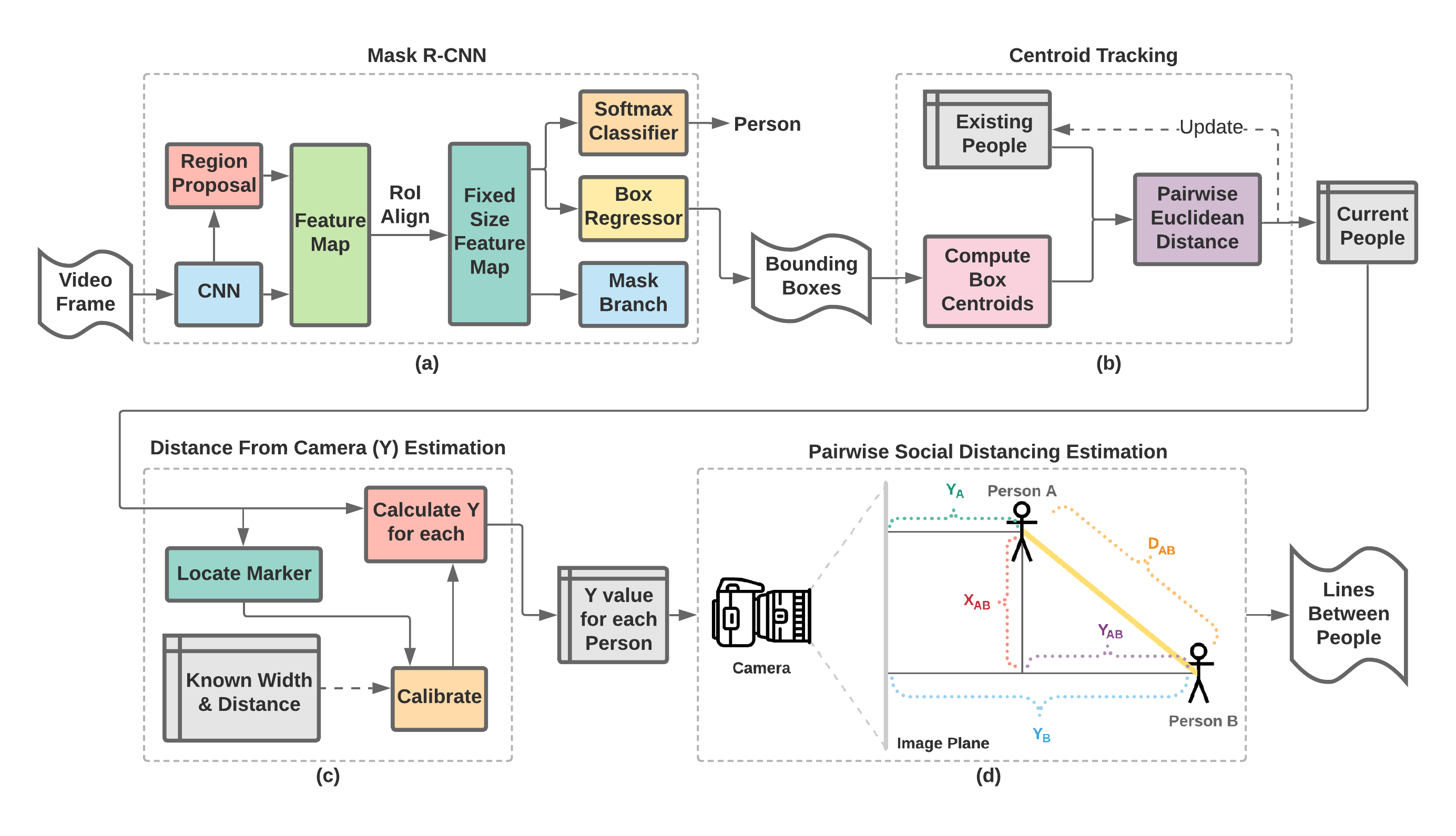}
    \caption{Workflow of SD-Measure Framework}
    \label{fig:sdworkflow}
    \end{minipage}%
\end{figure*}

This section elucidates a novel framework called SD-Measure which aims to determine whether a set of people are following `Social Distancing' guidelines of maintaining a minimum distance of 6 feet (or 1.8 metres) \cite{mref2} when observed from video footage of a public area. The proposed framework performs four major tasks in the following order:
\begin{enumerate}[label=\Alph*)]
    \item Person Detection
    \item Person Tracking
    \item Distance from Camera Estimation
    \item Pairwise Social Distancing Estimation
\end{enumerate}
The following sections explain the different stages involved in performing these tasks. The workflow for this proposed framework is illustrated in Figure \ref{fig:sdworkflow}.

\subsection{Person Detection}
We perform object detection using Mask R-CNN \cite{mref3} to detect people in a video frame. We use Mask R-CNN as it extends and improves Faster R-CNN \cite{mref4} by adding a mask branch and using Region of Interest (RoI) Align instead of RoI Pooling. RoI Align overcomes the issue of location misalignment existent in RoI Pooling. It achieves this by dividing the input proposals from the Region Proposal Network (RPN) into `bins' using bilinear interpolation. Apart from the improvement in the core accuracy over Faster R-CNN, Mask R-CNN is able to create masks for the detected object allowing the authorities to distinguish individual persons in a frame more easily.
\par In this application, the Mask R-CNN model takes 30 video frames per second and outputs a list containing coordinates of the bounding boxes, masks, and detection scores for each detected person. The given step is illustrated in Figure \ref{fig:sdworkflow} (a).

\subsection{Person Tracking}
After performing object detection, correctly detected people are retained based on their class IDs and detection scores after filtration from all the detected objects. This paves the way for the next task which is person tracking over the later time frames of the recording. Person tracking is accomplished by utilising the Centroid Tracking algorithm\cite{mref5}. This algorithm works by calculating the Euclidean separation between the centroids of identified people over successive frames as illustrated in Figure \ref{fig:sdworkflow} (b).\par

The primary premise of tracking people is to be able to consistently determine whether a pair of people are following social distancing or not even if the object detection algorithm is unable to detect the previously detected person in intermediate frames. This ensures that the proposed framework is robust against fluctuations in the output of the object detection model.\par

\subsection{Distance from Camera Estimation}
\label{sec:distancefromcam}
Once the people in the frame have been correctly identified and tracked, our goal is to estimate the real distance of each person from the camera in metres using an effective mathematical principle known as `Triangle Similarity'. This principle relies on a two-step process: (1) Focal length ($F$) determination using a located marker, and (2) Distance estimation of each person.\par

\subsubsection{Focal length ($F$) determination using a located marker}
This step uses a marker to calibrate and determine the focal length of the camera used in capturing the video frames. This is done by locating a marker object and  relating its perceived dimensions with its known dimensions. To achieve this, we first locate any person and denote it as the marker. Then, we calculate its width in pixels by taking the difference of the horizontal coordinates of the bounding box's top-left corner ($X_1$) and bottom-right corner ($X_2$) and denote it as $P$ as shown in Eq. \ref{eq:width}. 
\begin{equation}
\label{eq:width}
    P = X_2 - X_1
\end{equation}
We estimate this marker person to be at a distance $D$ from the camera either using its known distance or human perception. Then, we use a known value $W$ which is the width of the person in metres.\par
To actually determine the value of $W$, we determine the elbow-to-elbow distance of a person since elbows extend the furthest from the body when standing and facing the camera. To determine this distance, we consider the average waist size of the population from which the video footage originates and account for some buffer for the placement of the arms of the person around his/her body. In case the origin of the video footage is unknown, we use other features of the marker person such as race or gender to determine the average waist size.\par
Once the known width $W$, distance from camera $D$, and the width in pixels $P$ of the marker person have been tabulated, we calculate the focal length $F$ of the camera as shown in Eq. \ref{eq:focal}. 
\begin{equation}
\label{eq:focal}
    F = \frac{D \times P}{W}
\end{equation}

\subsubsection{Distance estimation of each person}
Once we have determined the focal length $F$ of the camera used for capturing the given video footage, we can estimate the real distance of a person from the camera $Y$ based on the width of its bounding box in pixels $P$ and the known width $W$ (as determined previously) using Eq. \ref{eq:distancefromcam}.\par
\begin{equation}
\label{eq:distancefromcam}
    Y = \frac{F \times W}{P}
\end{equation}
This distance from the camera $Y$ is subsequently calculated for every tracked person in the frame. The procedure for this step of the proposed framework is illustrated in Figure \ref{fig:sdworkflow} (c).\par

\subsection{Pairwise Social Distancing Estimation}
After we have estimated the real distance of each tracked person from the camera, our next objective is to determine the person-to-person distance between each each pairwise-distinct set of people. This equates to $_nC_2$ number of distinct pairs where $n$ is the number of tracked people. The following steps explain the calculation of the real distance between a pair of tracked people which is also illustrated in Figure \ref{fig:sdworkflow} (d). For explanation purposes, we'll denote the two people in a pair as Person $A$ and Person $B$.\par
Firstly, we retrieve the estimation of real distance calculated in the previous step and denote them as $Y_A$ and $Y_B$ for the two people respectively. Then, we calculate the absolute difference in the distance from the image plane between the two people and denote it as $Y_{AB}$ using Eq. \ref{eq:vertical}.
\begin{equation}
\label{eq:vertical}
Y_{AB} = |Y_B - Y_A|
\end{equation}
Secondly, we determine the horizontal distance between the two people by finding the absolute difference in the horizontal coordinates of their centroids, which are denoted as $X_A$ and $X_B$ respectively. The absolute difference is denoted as $X_{AB}$ and is calculated as shown by Eq. \ref{eq:horizontal}.
\begin{equation}
\label{eq:horizontal}
X_{AB} = |X_B - X_A|
\end{equation}
A key point to note is that the distance $Y_{AB}$ is calculated in metres whereas the distance $X_{AB}$ is calculated in pixels. Hence, the next imperative step is to convert the units of the distance $X_{AB}$ into metres.\par
\begin{figure}[H]
    \centering
    \includegraphics[width = 80mm, height = 45mm]{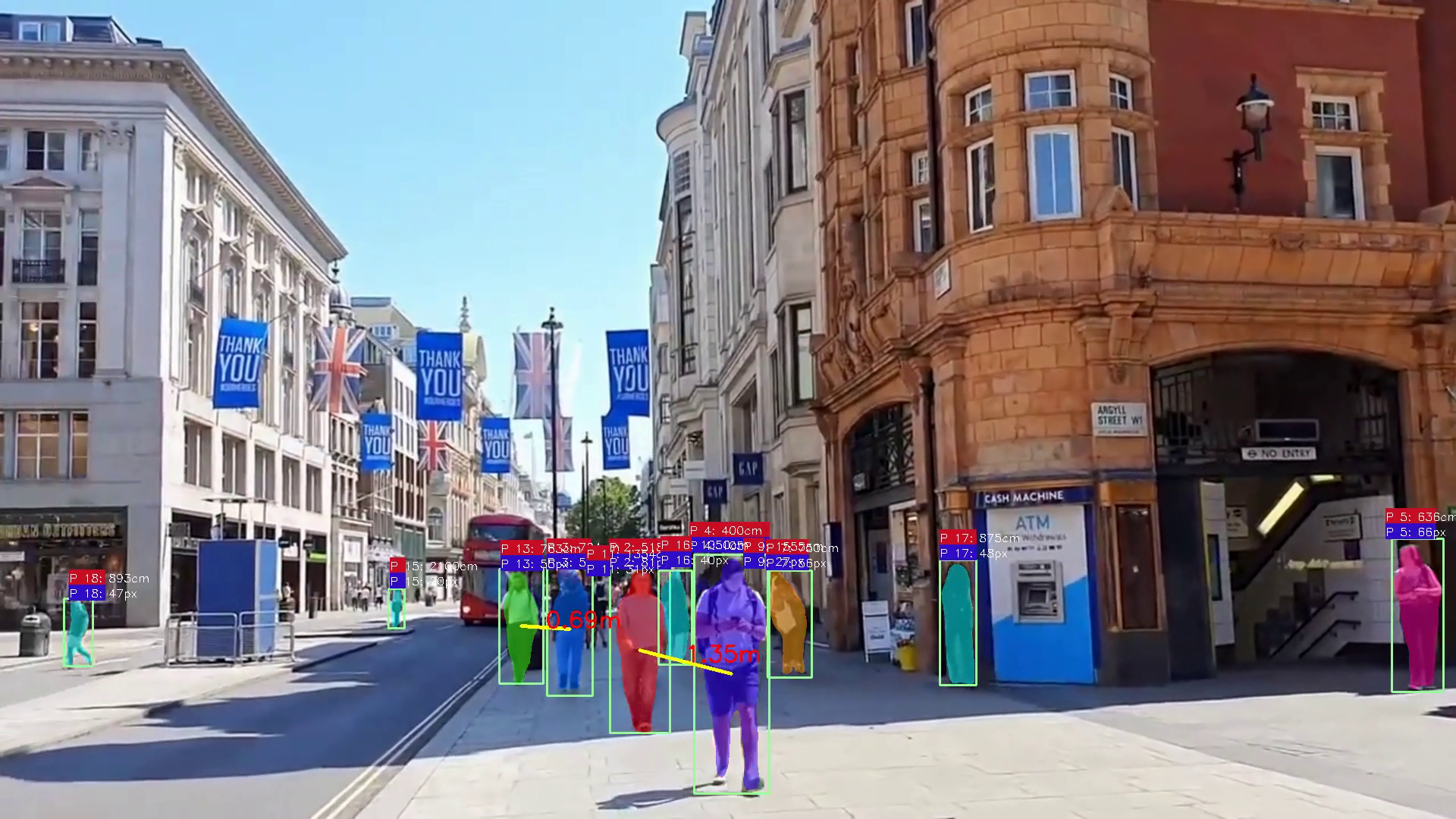}
    \includegraphics[width = 80mm, height = 45mm]{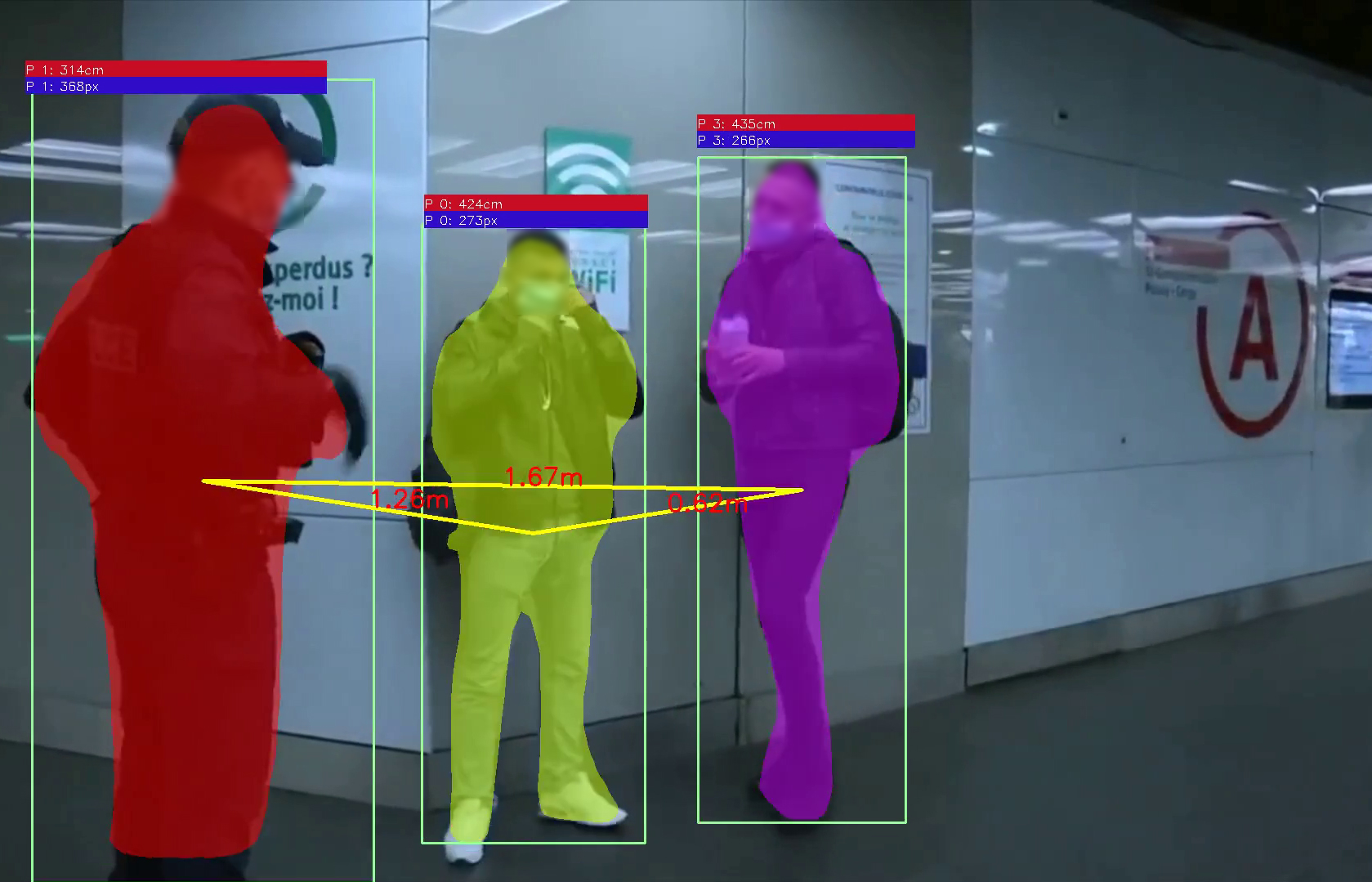}
    \caption{Visualisation of the results obtained by SD-Measure framework}
    \label{fig:results}
\end{figure}

To achieve this, we first calculate the widths of both Person $A$ and Person $B$ using Eq. \ref{eq:width} and denote them as $P_A$ and $P_B$ respectively. Then, we calculate the average of these widths $P_{AB}$ using Eq. \ref{eq:widthavg}.\par
\begin{equation}
\label{eq:widthavg}
P_{AB} = \frac{|P_A + P_B|}{2}
\end{equation}
We calculate this average width $P_{AB}$ in order to normalize the discrepancies in individual widths when determining our next measure which is `pixels per metre' denoted as $PPM$. We determine the $PPM$ based on the average width $P_{AB}$ and the known width $W$ (obtained from Section \ref{sec:distancefromcam}) using Eq.\ref{eq:ppm}.\par
\begin{equation}
\label{eq:ppm}
PPM = \frac{P_{AB}}{W}
\end{equation}
Once we have determined the $PPM$, we calculate the absolute difference in the horizontal distance in metres ${X'}_{AB}$ using Eq.\ref{eq:horizontalmetres}.\par
\begin{equation}
\label{eq:horizontalmetres}
{X'}_{AB} = \frac{X_{AB}}{PPM}
\end{equation}
After we have determined the absolute difference in distance from the camera $Y_{AB}$ in metres and the absolute difference in the horizontal distance ${X'}_{AB}$ in metres, we calculate the centroid-to-centroid distance in metres between Person $A$ and Person $B$. We use the Pythagorean Theorem to obtain the real distance $D_{AB}$ in metres as shown in Eq. \ref{eq:distanceD}.\par
\begin{equation}
\label{eq:distanceD}
D_{AB} = \sqrt {\left( {X'}_{AB} \right)^2 + \left( Y_{AB} \right)^2 }
\end{equation}
This is the distance in metres estimated between a pair of people to determine whether `Social Distancing' guidelines \cite{mref2} are being followed. If this distance is less than 1.8 metres, a line is drawn connecting the centroids of the two people in the pair to visualize people who are in violation of the given guidelines when observed from video footage of a public area. Visualization of the results obtained by the proposed approach is depicted in Figure \ref{fig:results}.\par

%
\section{Experimental Evaluation}
In this section, we discuss the dataset used for conducting this study and the results obtained by the proposed approach. The experiments were conducted on Google Colab \cite{refx1} with Intel(R) Xeon(R) 2.00 GHz CPU, NVIDIA Tesla T4 GPU, 16 GB GDDR6 VRAM and 13 GB RAM. All programs were written in \textit{Python} - 3.6 and utilised \textit{OpenCV} - 4.2.0, \textit{Keras} - 2.3.0 and \textit{TensorFlow} - 2.2.0.\par

\subsection{Dataset Used}
\label{dataset}
\begin{figure}[!ht]
    \centering
    \includegraphics[width=80mm, height=80mm]{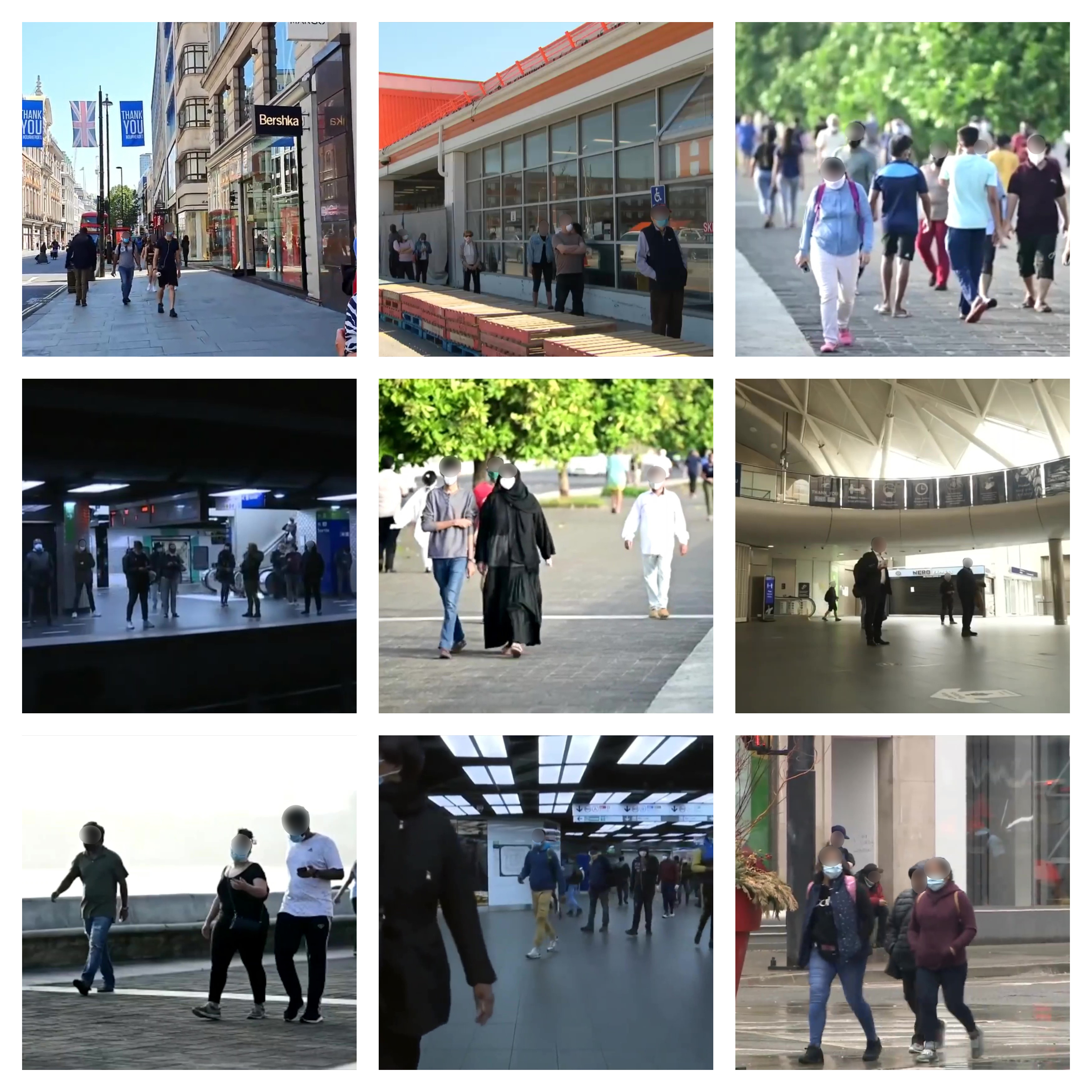}
    \caption{Video samples from the CVFD dataset}
    \label{fig:dataset}
\end{figure}

\begin{figure}[!ht]
    \centering
    \includegraphics[width=90mm, height=110mm]{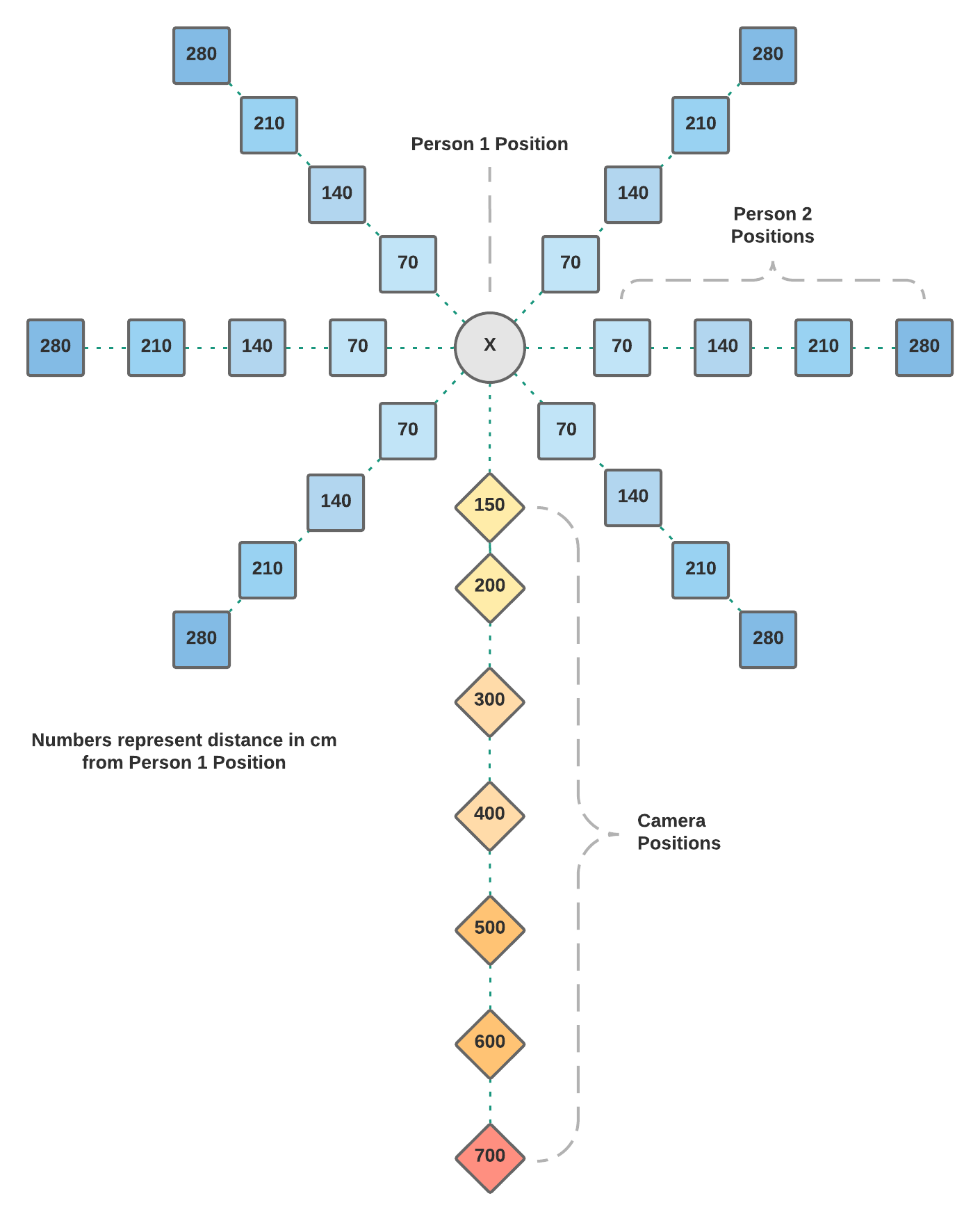}
    \caption{Layout for CPID dataset}
    \label{fig:positions}
\end{figure}

The following datasets were used in the development of the proposed framework:
\begin{enumerate}
    \item MS COCO: The Microsoft Common Objects in Context (MS COCO) dataset \cite{mref6} is a well-known dataset that has annotations for instance segmentation and bounding boxes used to evaluate how well object detection and image segmentation models perform. The dataset has been utilized to perform training on the Mask R-CNN \cite{mref3} framework on the `person' class.
    \item Custom Video Footages Dataset (CVFD): This comprises of a collection of footages of public places from multiple geographical locations, compiled from YouTube. The videos capture the movement of people in public areas after the imposition of various safety rules and regulations in wake of the COVID-19 pandemic. This dataset contains videos having different camera angles, varying illumination conditions, noise, and an average frames per second (FPS) of 30. Figure \ref{fig:dataset} illustrates a few sample videos present in this dataset.
    \item Custom Personal Images Dataset (CPID): We created our own dataset by taking images of two people, namely Person 1 and Person 2. Person 1 was used as a reference point and Person 2 was relocated in fixed positions with known distance from the reference point. The distance between the reference point and the camera was also known and varied at fixed distances. The primary object of this dataset was to determine the efficacy of the proposed framework in determining the true distance between people. The various fixed positions and distances from the camera with respect to the reference position is illustrated in Figure \ref{fig:positions}.
\end{enumerate}
\subsection{Experimental Results and Statistics}
We have evaluated the proposed approach in three parts: person detection using Mask R-CNN, detection of social distancing lines, and the efficacy of the measure of the distance between a pair of people.\par
\begin{gather}
    \textit{Precision} = \frac{TP}{TP+FP} \times 100\% \label{eq:prec}\\   
    \textit{Recall} = \frac{TP}{TP+FN} \times 100\% \label{eq:recall}\\
    \textit{False Alarm Rate} = \frac{FP}{TN+FP}\times 100\% \label{eq:FAR}\\
    \textit{Accuracy} = \frac{TP+TN}{TP+TN+FP+FN}\times 100\% \label{eq:accuracy}\\
    \textit{Standard deviation (S)} = \sqrt{\frac{\sum(x_i - \bar{x})^2}{n-1}}
    \label{eq:stdev}\\
    \textit{\% error} = \frac{|true\: distance - estimated\: distance|}{true\: distance}\times 100\%
    \label{eq:error}
\end{gather}
We estimate the performance of the person detection using Mask R-CNN \cite{mref3} based on two parameters: precision (Eq. \ref{eq:prec}), and recall (Eq. \ref{eq:recall}), which are calculated from the number of True Positives (TP), False Positives (FP), and False Negatives (FN). These parameters have been determined based on the results when tested on the CVFD dataset (as explained in Section \ref{dataset}) and are shown in Table \ref{tb:mrcnnres}.\par
{\renewcommand{\arraystretch}{1.5}%
\begin{table}
\centering
\begin{tabular}{|c|c|}
\hline
   Precision & 97.44\%  \\ \hline
   Recall & 90.39\% \\ \hline
\end{tabular}
\caption{\label{tb:mrcnnres} Evaluation result for Mask R-CNN tested on CVFD dataset}
\end{table}
} 
Since every part of the image where we do not predict a person is considered a negative, the measurement of True Negative is considered ineffectual. The focus of the person detection algorithm is to detect people with a high precision even with a trade-off of a slightly reduced recall. This is because we are using a tracking algorithm which can track people over several frames to allow the detection algorithm to have false negatives without affecting the efficacy of the overall framework. On the contrary, a lower precision would translate to a greater number of false positives which would lead to incorrect detections being tracked for several frames and affecting the effectiveness of the framework.\par
We evaluated the detection of social distancing lines based on four parameters: precision, recall, false alarm rate (Eq. \ref{eq:FAR}), and accuracy (Eq. \ref{eq:accuracy}). As explained in Section \ref{approach}, we draw a line between a pair of people if the distance between them has been estimated to be less than 1.8 metres. We analyzed each frame of the CVFD dataset (refer to Section \ref{dataset}) and determined whether a line has been correctly drawn or not for each pair-wise distinct set of people. We considered a True Positive (TP) when a line was perceived to be correctly drawn, a False Positive (FP) when a line was incorrectly drawn, a True Negative (TN) when a line was correctly not drawn, and a False Negative (FN) when a line was incorrectly not drawn for a pair of people. Since it is not possible to identify the actual distance between each pair of people from a video footage, we used human perception versus the detections made by the proposed framework to determine the aforementioned values. To ensure robustness and to minimize errors due to bias and other random errors, we used a majority voting system among the five authors to determine whether a line should be drawn or not between each pair of people in a given frame and then compared with the output of the proposed framework. The results have been tabulated in Table \ref{tb:socialdistancing}.\par
{\renewcommand{\arraystretch}{1.5}%
\begin{table}
\centering
\begin{tabular}{|c|c|}
\hline
   Precision & 86.06\%  \\ \hline
   Recall & 83.99\% \\ \hline
   False Alarm Rate & 3.27\% \\ \hline
   Accuracy & 94.26\% \\ \hline
\end{tabular}
\caption{\label{tb:socialdistancing} Evaluation result for proposed framework for social distancing lines drawn when tested on CVFD dataset}
\end{table}
}
We evaluated the efficacy of the measure of the distance between a pair of people based on two statistical parameters: standard deviation (Eq. \ref{eq:stdev}) and percent error (Eq. \ref{eq:error}). Since the proposed framework estimates the actual distance between each pair of people (as explained in Section \ref{approach}), we used statistics to evaluate the output of the proposed framework when tested on CPID dataset (refer to Section \ref{dataset}). 
\par
\begin{figure}[!ht]
    \centering
    \includegraphics[width=90mm, height=55mm]{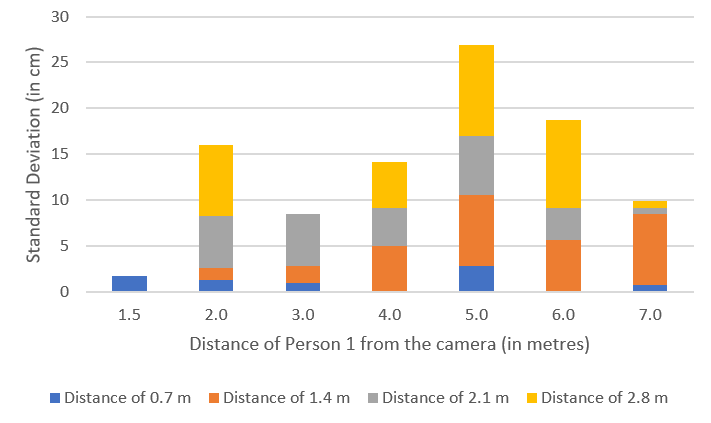}
    \caption{Standard deviation of the output of the proposed framework when tested on the CPID dataset. The key represents the distance between Person 1 and Person 2 in the given layout.}
    \label{fig:stdev}
\end{figure}

 We used standard deviation to determine the framework's precision, which is a measure of how close the measurements are with each other. We calculated the standard deviation for various fixed distances between Person 1 and Person 2 at each of the fixed distances between the camera and Person 1 in the layout which is shown in Figure \ref{fig:positions}. The standard deviation values have been represented as stacks for each distance between the camera and Person 1 as shown in Figure \ref{fig:stdev}. It can be observed that the proposed approach shows greater precision in the estimated distance when the reference person is at a close-medium distance from the camera.
\par
\begin{figure}[!ht]
    \centering
    \includegraphics[width=90mm, height=55mm]{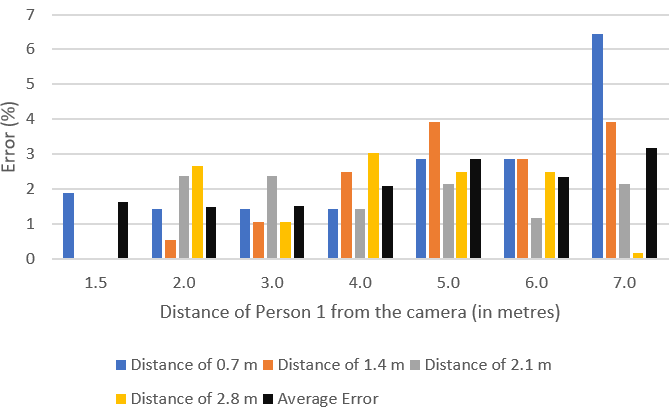}
    \caption{Percent error of the output of the proposed framework when tested on the CPID dataset. The key represents the distance between Person 1 and Person 2 in the given layout.}
    \label{fig:error}
\end{figure}

 Similarly, we used percent error to determine the framework's accuracy, which is a measure of how close the measurements are to the accepted or true value. We calculated the percent error for various fixed distances between Person 1 and Person 2 at each of the fixed distances between the camera and Person 1 and is shown in Figure \ref{fig:error}. We have also represented the average error for each distance between Person 1 and the camera as well. It can be observed that as the distance between the camera and the reference person (i.e. Person 1) is increased, the error increases. This demonstrates that the proposed approach has a higher accuracy for determining the distance between people who are located at a closer range with respect to the camera.

\section{Conclusion and Future Works}
\label{conclusion}
In this work, a novel framework for detecting social distancing from video footage is proposed. By leveraging state-of-the-art object detection models for identifying people followed by an authentic distance from camera and a pairwise distance estimation algorithms, we deduce whether social distancing guidelines are being followed. The resulting approach is effective and was evaluated on a custom video footages dataset obtained for this research, as evidenced by the high accuracy value along with good precision and recall values. The low false alarm rate further validates the potency of the proposed approach. The approach can be fine-tuned for better performance according to the specific environment in consideration. In addition, large obstacles obstructing the field of view of the cameras may affect the tracking of people and in turn correct estimation of social distancing, which can be addressed in future work.

\section{Acknowledgement}
We thank Google Colaboratory for providing the necessary computational resources for conducting the experiments and YouTube for availing the videos used in this dataset. We also thank our institute, the National Institute of Technology Warangal for its constant support and encouragement to undertake research.
%
\bibliographystyle{IEEEtran}
\bibliography{main}
\end{document}